\documentclass[conference,a4paper]{IEEEtran}
\usepackage{graphicx, epstopdf}
\usepackage{amsmath,amssymb} 

\usepackage{algorithm}
\usepackage{algorithmic}

\usepackage{color}
\usepackage{soul}

\usepackage{subfig}
\usepackage{hyperref}

\usepackage{multirow}
\usepackage{tabularx} 
\usepackage{booktabs}

\newcolumntype{L}[1]{>{\hsize=#1\hsize\raggedright\arraybackslash}X}%
\newcolumntype{R}[1]{>{\hsize=#1\hsize\raggedleft\arraybackslash}X}%
\newcolumntype{C}[1]{>{\hsize=#1\hsize\centering\arraybackslash}X}%

\graphicspath{{figures/}}

\newcommand{\etal}{\textit{et al.}}

\newcommand{\cG}{\mathcal{G}}
\newcommand{\cE}{\mathcal{E}}
\newcommand{\cV}{\mathcal{V}}

\newcommand{\cC}{\mathcal{C}}
\newcommand{\cR}{\mathcal{R}}
\newcommand{\cF}{\mathcal{F}}

\newcommand{\cX}{\mathcal{X}}
\newcommand{\cY}{\mathcal{Y}}

\newcommand{\bT}{\mathbf{T}}
\newcommand{\bR}{\mathbf{R}}
\newcommand{\bX}{\mathbf{X}}

\newcommand{\bt}{\mathbf{t}}
\newcommand{\bx}{\mathbf{x}}

\newcommand{\bE}{\mathbf{E}}

\ifCLASSINFOpdf
\else
\fi
\begin{document}
%
\title{Monocular Rotational Odometry with Incremental Rotation Averaging and Loop Closure}

\author{\IEEEauthorblockN{Chee-Kheng Chng,   Alvaro Parra,     Tat-Jun Chin, Yasir Latif}
\IEEEauthorblockN{School of Computer Science, University of Adelaide}
}

\maketitle

\begin{abstract}
Estimating absolute camera  orientations is essential for attitude estimation tasks. An established approach is to first carry out visual odometry (VO) or visual SLAM (V-SLAM), and retrieve the camera orientations (3 DOF) from the camera poses (6 DOF) estimated by VO or V-SLAM. One drawback of this approach, besides the redundancy in estimating full 6 DOF camera poses, is the dependency on estimating a  map (3D scene points) jointly with the 6 DOF poses due to the basic constraint on structure-and-motion. To simplify the task of absolute orientation estimation, we formulate the monocular rotational odometry problem and devise a fast algorithm to accurately estimate camera orientations with 2D-2D feature matches alone. Underpinning our system is a new incremental rotation averaging method for fast and constant time iterative updating.  Furthermore, our system maintains a view-graph that 1) allows solving loop closure to remove camera orientation drift, and 2) can be used to warm start a V-SLAM system. We conduct extensive quantitative experiments on real-world datasets to demonstrate the accuracy of our incremental camera orientation solver. Finally, we showcase the benefit of our algorithm to V-SLAM: 1) solving the known rotation problem to estimate the trajectory of the camera and the surrounding map, and 2) enabling V-SLAM systems to track pure rotational motions.




%


\end{abstract}



%
\IEEEpeerreviewmaketitle

\section{Introduction}
Visual odometry~\cite{engel2017direct, forster2016svo, zhan2019visual} (VO) and visual SLAM~\cite{mur2017orb, engel2014lsd} (V-SLAM) estimate the poses of a moving camera from the captured sequence of images, which is relevant to many real-world technologies such as autonomous driving, UAV, and virtual-reality applications. In particular, \emph{monocular} VO/V-SLAM is of high interest due to its low cost, low power consumption, and its easy-to-setup nature. 

Given a set of images within a captured sequence, $\{I_j\}$, monocular VO and V-SLAM systems find the pose
\begin{align}
    \bT_j = \begin{bmatrix}
				\bR_j & \bt_j \\
				0 & 1\\
		\end{bmatrix}
\end{align}
\noindent where $\bR_j \in SO(3)$ is the camera orientation represented by rotation matrix, and $\bt_j\in\mathbb{R}^3$ is its 3D position in a common coordinate system. Apart from estimating the camera pose, V-SLAM systems also maintains (and output) a \emph{global map} with the coordinates of the observed scene points $\cX = \{X_i \;|\; X \in \mathbb{R}^3\}$ which can be used to 1) relocalise the camera when it lost track, and 2) perform loop closure. 

There exist some applications when the camera orientation is the sole interest, \emph{attitude estimation}~\cite{crassidis2007survey, khosravian2017discrete}, and the \emph{known rotation problem}~\cite{sim2006recovering, bustos2019visual} are among the well studied ones. Attitude estimation predicts the orientation of a moving vehicle based on high-end IMUs. Motivated by the lower cost aspects (in terms of both power consumption and price), Khosravian \etal~\cite{khosravian2017discrete} introduced an algorithm to incorporate GPU velocity readings to refine VO's orientation estimates. Secondly, in the V-SLAM domain, given the orientation priors, the known rotation problem~\cite{sim2006recovering} can be solved in a quasi-convex framework to obtain the camera position and the observed 3D scene points. We argue that VO and V-SLAM are not the best choices for these applications demanding camera orientation computation alone. Instead, we propose a novel monocular rotational odometry system to estimate absolute camera orientations from relative camera orientations under \emph{epipolar geometry}~\cite{hartley2003multiple} for which 2D-2D feature correspondences is sufficient. Underpinning our system is a proposed incremental rotation averaging method, which is fast and accurate.

Apart from the system-level advantages, such as a smaller memory footprint, our proposed system exhibits two inherent strengths over VO/V-SLAM systems. Firstly, our method has no requirement in creating and maintaining a local map which is essential for VO and V-SLAM systems. More specifically, VO systems solve PnP~\cite{lepetit2009epnp} which takes 2D-3D feature correspondences to jointly estimate the camera orientation and translation; while V-SLAM systems perform local BA\cite{engels2006bundle} to jointly optimise the camera poses and scene points. The requirement of the 3D scene points mainly stems the translation of the camera, which is unnecessary for our objective. 

Secondly, maintaining a 3D map is fundamentally unfeasible during periods of pure rotation motion~\cite{hartley2003multiple}. It is geometrically impossible to triangulate the depths of 3D scene points at the limit of cameras approaching no parallax. During pure rotation motion, only a panoroma map (with no depth information) can be created for the observed features~\cite{pirchheim2013handling, gauglitz2012live}. However, panoramas can not be used together with the  3D map in the joint estimation frameworks such as PnP and BA. As a partial remedy, most VO/V-SLAM systems rely on some form of model selection~\cite{torr1998maintaining, gauglitz2012live} to invoke different routines depending on the type of motion (standard rigid motion and pure rotation motion). Without the need of maintaining a map, our proposed method leverages a \emph{motion robust} relative pose estimation algorithm~\cite{kneip2013direct}, which allows us to estimate the relative orientation in a unified framework.

Our proposed method relates to VO systems in the absence of global mapping. Since the incapacity of globally constraining the camera pose, most of VO systems advocate on reducing frame-to-frame drift. For this reason, recent active research in the field focus on improving front-end components such as feature descriptors~\cite{agrawal2015learning}, feature correspondences selection~\cite{zhan2019visual}, and accurate single-view depth estimation \cite{ummenhofer2017demon}. However, all of the mentioned algorithms involve deep networks which usually come at the expense of processing time, computation power, and training data. On the contrary, our proposed method uses ORB descriptors~\cite{rublee2011orb} (the fastest but less robust) and focuses on refining the estimates of camera orientation with \emph{rotation averaging}~\cite{chatterjee2013efficient,hartley2013rotation} formulated in an incremental fashion (detailed in Sec.~\ref{sec:rel_work:rot_avg}).

From our empirical evidence over long sequences in the KITTI dataset~\cite{geiger2012we}, incremental rotation averaging alone is insufficient to combat drift accumulation. Hence, we integrated a loop closure component in our system, which invokes global rotation averaging routine to distribute the accumulated drift to the node in the view-graph upon detecting loop-closure. The loop-closure module is built on top of the appearance based loop-closure method in  ORB-SLAM2~\cite{mur2017orb}. Due to the lack of global mapping, we replaced the point clouds based geometrical verification process with a 2D-2D feature matches based algorithm (see Sec.\ref{sec:loop_closure}). In the aspect of having a loop-closure module, our system is related to V-SLAM systems. Hence, we compare our method to the state-of-the-art monocular V-SLAM system, ORB-SLAM2\cite{mur2017orb}, over two real-world datasets and demonstrate that our proposed method achieves competitive results.


We summarise our contributions as
\begin{enumerate}
    \item An efficient incremental rotation averaging method\footnote{C++ implementation will be available.}. 
    \item A visual rotational odometry system which is robust against any camera motion (Sec.~\ref{sec:pipeline:rel_rotation}), capable of loop-closing (Sec.~\ref{sec:loop_closure}), and that achieves state-of-the-art accuracy on real-world datasets (Sec.~\ref{sec:exp:accu}).
\end{enumerate}

\section{Related Works}

\subsection{Monocular Visual Odometry and Visual SLAM systems}
Our approach is closer to VO than V-SLAM as we do not require computing and maintaining a global map. The interested reader can find a comprehensive survey for VO in~\cite{scaramuzza2011visual}. In essence,  conventional VO pipelines first identify feature matches, obtain 2D-3D correspondences after triangulating the scene map, and estimate relative motion between consecutive frames. Finally, they estimate the absolute poses of the camera by chaining the relative motions. An emerging trend in VO systems is the incorporation of deep networks to learn and predict 1) feature correspondences with an optical flow network~\cite{zhan2019visual}, 2) depths of the observed features with a single image~\cite{yin2018geonet}, and 3) camera poses~\cite{zhou2017unsupervised}. Since conventional VO algorithms do not maintain a global map and do not perform loop closure, they focus on reducing the drifts between each pose estimate with better front-end inputs (i.e., feature correspondences, accurate depth predictions). On the other hand, our proposition uses the fastest but less robust ORB-descriptors for the 2D-2D feature matching process, and we use our proposed incremental and robust rotation averaging algorithm to produce accurate orientation estimates. 



In addition to estimating the camera pose, monocular V-SLAM systems also estimate the observed 3D scene points (known as \emph{mapping}). In a broad sense, V-SLAM methods can be classified into three groups: indirect-based, direct-based, and the hybrid group. ORB-SLAM2~\cite{mur2017orb} is the state-of-the-art system in the indirect-based category. ORB-SLAM2 operates as follows. It first extracts ORB image features and identifies feature matches to previous images. To jointly estimate camera poses and the coordinates of scence points, ORB-SLAM2 leverages in a combination of well-studied geometry algorithms (including relative motion estimation and PnP) with large scale non-linear optimisation frameworks (e.g., pose graph optimisation and bundle adjustment). Engel \etal~\cite{engel2017direct} propose a direct based system that utilises all the pixel information instead of selecting feature correspondences and minimises the \emph{photometric loss} instead of the reprojection error as in conventional indirect methods. The system, namely DSO, was later extended with loop-closure in \cite{gao2018ldso}. Our work is closely related V-SLAM works in the loop closure aspect, and specifically close to ORB-SLAM2 as our system extracts and matches ORB-features, and performs loop-closure with DBoW2 featuress~\cite{galvez2012bags}.

\subsection{The View-Graph}
Conventional VO and V-SLAM systems maintains a so-called \emph{view-graph} $\cG = (\cV,\cE)$ with edges $(j,k)\in \cE$ connecting images $I_j,I_k$ for which a relative motion can be estimated. Typically, the view-graph relates the image pairs capturing overlapped scenes such that sufficient epipolar constraints exist to obtain the essential matrix $\bE_{j,k}$. In our system, the nodes of the view-graph relate to camera orientations, and the edges to the relative orientation between the connected views.

\subsection{Rotation Averaging} \label{sec:rel_work:rot_avg}

In the absence of noise and outliers, the absolute orientation $\bR_k$ for an image $I_k$ can be directly estimated by chaining the orientation 
$\bR_{j}$ of another image $I_j$ to the relative orientation $\bR_{j,k}$ between $I_j$ and $I_k$, for which
\begin{align} \label{eqn:rel_rot}
\bR_k = \bR_j \bR_{j,k}
\end{align}
holds. Thus, under those ideal conditions, we could constraint $\cG$ to be a ``loop'' (only connecting temporary adjacent images) and obtain all absolute orientations (up to some arbitrary orientations) by simply chaining them by applying Eq.~\eqref{eqn:rel_rot}. However, in the VO/V-SLAM application settings where the presence of outliers and noise is inevitable, estimating orientations from $\cG$ being a loop will inevitably produce inaccurate results as this setting requires that all relative orientations are precisely computed. Instead, \emph{rotation averaging} solves for the absolute orientations by minimising the discrepancies between relative and absolute orientations across all edges in $\cG$
\begin{align} \label{eqn:rot_avg}
\min_{\{\bR_j\}} \sum_{(j,k)\in \cE} \rho ( d( \bR_{j,k}, \bR_k \bR_j^{-1}) ),
\end{align}
where $\rho$ is a loss function (e.g. $\rho(x) = x^2$) and $d: SO(3)\times SO(3) \mapsto \mathbb{R}_+$ is a distance between rotation matrices (e.g., the geodesic distance in SO(3)). 



Rotation averaging~\eqref{eqn:rot_avg} was first introduced by Govindu~\cite{govindu2004lie}, and has since been further explored by numerous works~\cite{hartley2013rotation, chatterjee2013efficient, eriksson2018rotation}. 
Chatterjee \etal ~\cite{chatterjee2013efficient} proposes an efficient and robust algorithm (L1-IRLS) to solve~\eqref{eqn:rot_avg}. L1-IRLS is an iterative reweighted least-squares algorithm which finds a local optimum in the Lie group SO(3) with a robust initialisation scheme. Specifically, the initial estimates are obtained by solving an approximated algorithm based on an easy to solve yet robust $\ell_1$-based optimisation problem. As default, L1-IRLS takes $\rho$ as the Huber-like robust loss function and solve of the weights (of the relative rotations) and the absolute rotations alternatively until convergence. Our proposed incremental rotation averaging method (see Sec.\ref{sec:pipeline:IRA}) is an extension to L1-IRLS.

Rotation averaging is relevant in the V-SLAM field. Carlone \etal~\cite{carlone2015initialization} propose to first estimate the orientation components to initialise the subsequent pose-graph optimisation. Parra \etal~\cite{bustos2019visual} propose an L-infinity V-SLAM framework that detaches the camera orientation estimates from the translation component and the observed scene points. Although both of the mentioned works use rotation averaging to estimate the absolute orientation of each camera pose, none of them formulates and solve the problem incrementally.




\subsection{Relative Orientation Estimation} \label{sec:rel_work:rel_rot_est}

For every incoming frame, we first identify the 2D-2D feature matches with the previously processed frames within a fixed window (more details in Sec.~\ref{sec:pipeline:rel_rotation}), and estimate their relative orientations. If there are sufficient feature matches, we add a new node to the view-graph and the edges related to the estimated relative orientations. 

The \emph{five-points algorithm}~\cite{nister2004efficient} is the \emph{``go-to''} method in estimating the relative motion between a pair of frames. However, the method breaks down when the baseline between the input frames vanishes, resulting in undefined essential matrix~\cite{hartley2003multiple}. One real-world scenario of the mentioned degeneracy is the pure rotation motion (or rotation-only motion) where the camera is rotating around a principle axis and not moving in the 3D euclidean space. Hence, VO/V-SLAM systems usually incorporate model selection techniques (stemming from~\cite{torr1998maintaining, gauglitz2012live}) to select either essential matrix or homography matrix (which is defined in pure rotation motion) is appropriate to model the current motion. Instead of relying on model selection, we adopted the relative orientation estimation algorithm in~\cite{kneip2013direct} which is robust against any motion. Underpinning Kneip \etal's method is the \emph{normal epipolar constraint} (first introduced in~\cite{kneip2012finding}), a novel epipolar geometry constraint capable of decoupling the relative orientation and the translation direction between a pair of frames. As such, the constraint is well defined under both standard motion and rotation-only motion.

\section{Methods}
\begin{figure}
	\centering
	{\includegraphics[width=0.4\textwidth]{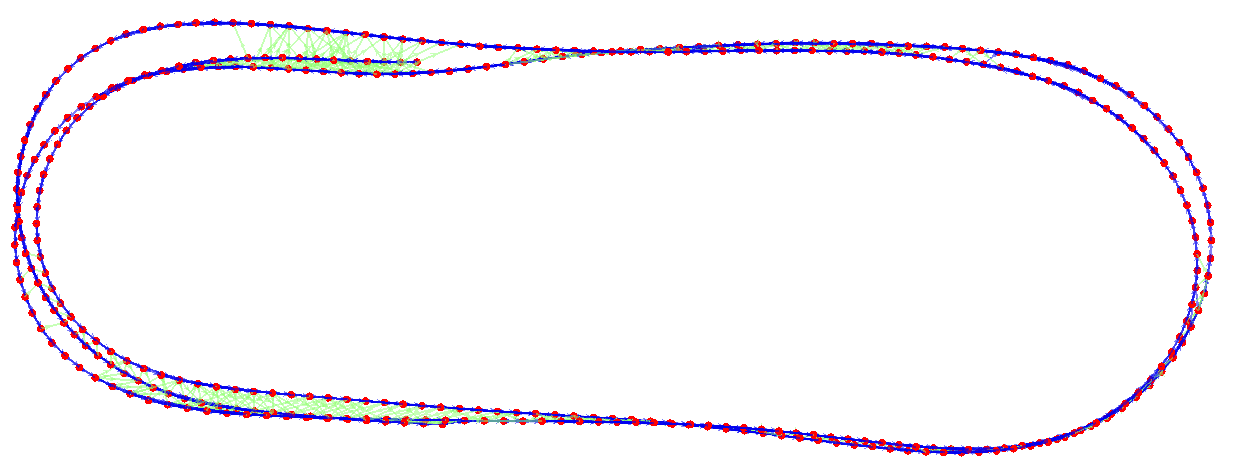}} \hspace{3em}
	\caption{View-graph of our system for the Carpark dataset. Nodes (in red) are at the camera locations. Edges represent covisibility. Loop-closure edges are highlighted in green.}\label{fig:mt_vg}
\end{figure}

Alg.~\ref{alg:pome} describes our system which advocates to keep updated a view-graph as the camera captures new frames. Contrasted to V-SLAM, our view-graph relates absolute  (nodes) and relative (edges) orientations alone (i.e., without the translation components as in V-SLAM systems). Fig.~\ref{fig:mt_vg} depicts the final view-graph for the Carpark dataset (see Sec.~\ref{sec:exp:datasets_metrics}) where each node is drawn with their 3D position we obtained from the ground-truths.

The core optimisation component in our method is the incremental rotation averaging routine (Alg.\ref{alg:pome}, Line~\ref{alg:pome:line:inc_rot_avg}) which solves a new rotation averaging formulation (Sec.~\ref{sec:pipeline:IRA}) to anchor the solution within a window of orientations $\cR_{\text{window}}$ to the previously estimated orientations out of the window.


\begin{algorithm}
\begin{algorithmic}[1]
    \STATE \textbf{global variables:} View-graph $\cG = (\cV, \cE)$. 
	\FOR {each new frame $k = 1, 2, \ldots$}
	\STATE $f \leftarrow k - |\cF_{\text{window}}|$. \label{alg:pome:line:front_end_start}
	
	\STATE  $\cY_k \leftarrow \text{feature\_extraction}(I_k)$. \label{alg:pome:line:feature_extraction}
	
	\FOR {j = f,\ldots ,k-1}
	
	    \STATE $\cC_{j,k} \leftarrow \text{feature\_matching}(\cY_j, \cY_k)$. \label{alg:pome:line:feature_matching}
	
	    \STATE ($\bR_{j,k}, \cC_{j,k}')\leftarrow \text{relative\_rotation}(\cC_{j,k})$. \label{alg:pome:line:rel_rotation}
	    
	    \IF {$| C_{j,k}' | > \theta_{\text{matches}}$}
    	    \STATE $\cG \leftarrow \text{upgrade\_view-graph}( \bR_{j,k} )$. \label{alg:pome:upgade_vg}
	    \ENDIF
	\ENDFOR
	
	\STATE Obtain $\cG' = (\cV', \cE')$ for $\cR_{\text{window}}$. \label{alg:pome:line:sub_graph}
	
	\STATE $\{\bR_j\}_{j \in \cE'} \leftarrow \text{inc\_rotation\_averaging}(\cG')$. \label{alg:pome:line:inc_rot_avg}
	
	\IF {loop closure detected} \label{alg:lc_begin} \label{alg:pome:lc_detec}
		\STATE $\{{\bR_j} \}_{j\in\cE} \leftarrow \text{glob\_rotation\_averaging}(\cG)$. \label{alg:pome:line:global_rot_avg}
	\ENDIF \label{alg:lc_end}
	\ENDFOR
\end{algorithmic}
\caption{Visual Rotational Odometry}
\label{alg:pome}
\end{algorithm}

\subsection{Feature Extraction and Matching}
\label{sec:pipeline:features_e_and_m}
We use ORB descriptors~\cite{rublee2011orb} as image features due to its superior speed performance. We adopted identical extraction technique as ORB-SLAM2~\cite{mur2017orb}. Upon extracting a set of ORB features $\cY_k$ in the current frame $I_k$ (Alg.~\ref{alg:pome}, Line~\ref{alg:pome:line:feature_extraction}), our system performs a local search (Alg.~\ref{alg:pome}, Line~\ref{alg:pome:line:feature_matching}) to find matches $\cC_{j,k} = \{ (\bx_i, \bx'_i)\;|\; \bx_i \in I_j \text{ and }  \bx'_i \in I_k \}$ in the previous frames ${\{I_j\}}_{j = f}^{k-1}$, where index $f$ is the first frame within a fixed window $\cF_{\text{window}}$. The local search algorithm assumes a high frame rate\footnote{which is a standard-setting in video sequences.}. Hence, the detected features remain close (in pixel coordinates) to their corresponding features in the neighbouring frames. 

\subsection{Motion Robust Relative Orientation Estimation}
\label{sec:pipeline:rel_rotation}
The conventional estimation of the relative orientation, $\bR_{j,k}$, (assuming known camera calibration) from decomposing the essential matrix
\begin{align}
 E_{j,k} = [\bt_{j,k}]_{\times} \bR_{j,k},    
 \end{align}
(where $[\bt_{j,k}]_{\times}$ is the skew-symmetric matrix representation of the cross product operator) is undefined for pure rotation motion ($\bt_{j,k} = \mathbf{0}$). Instead, we incorporate in our system (Alg.~\ref{alg:pome}, Line~\ref{alg:pome:line:rel_rotation}) the Kneip's method~\cite{kneip2013direct} which estimates the relative orientation independently of the translation. Thus, Kneip's method can correctly estimate $\bR_{j,k}$ even if $\bt_{j,k} = \mathbf{0}$. 

For an image point $\bx$, we define its \emph{bearing vector} 
\begin{align}
\mathbf{f} := \frac{\hat{\bx}}{\parallel \hat{\bx} \parallel},
\end{align}
where $\hat{\bx}$ is ${\bx}$ in normalised homogeneous coordinates (assuming known camera intrinsics). 

Kneip's method solves an iterative eigenvalue rank minimisation problem over a set of normal vectors
\begin{align} \label{eqn:normal_vecs}
\mathbf{n}_{i} := \mathbf{f}'_{i} \times \bR_{j,k} \mathbf{f}_{i}
\end{align}
of the \emph{epipolar plane}, where $(\mathbf{f}_i, \mathbf{f}'_i)$ are corresponding bearing vectors such that $(\bx_i, \bx'_i) \in \cC_{j,k}$. The essential geometry constraint (which is independent of the motion) is that all normal vectors must lie on the same plane. 

The eigenvalue rank minimisation problem finds the relative orientation that minimises the smallest eigenvalue $\lambda_{\mathbf{M},\text{min}}$
\begin{align} \label{eqn:eig_min}
\min_{\bR_{j,k}} \; \lambda_{\mathbf{M},\text{min}}
\end{align} 
of $\mathbf{M} = \mathbf{NN}^T$, the covariance matrix of the normal vectors $\mathbf{N} = [\mathbf{n}_1,\ldots, \mathbf{n}_n]$, 
 
To solve~\eqref{eqn:eig_min}, we initialise $\bR_{j,k}$ as $\bR_{j-1,k-1}$ assuming the rotation motion between neighbouring pair of frames should be close to each other\footnote{Similar to ORB-SLAM's constant velocity assumption which initialises the latest frame's camera pose with previously optimised poses}. 

Since $\mathbf{M}$ is a real symmetric and positive definite matrix with rank at most 2 (based on the coplanarity constraint on the normal vectors), it is, therefore, equivalent to an iterative rank minimisation. The optimisation framework is solved in a Levenberg-Marquardt scheme as detailed in~\cite{kneip2013direct}. Similar to the \emph{five-point method}~\cite{nister2004efficient}, Kneip's method is embedded in a RANSAC framework to remove outlying feature correspondences ($\cC_{j,k}' \subseteq \cC_{j,k}$ are the correspondences after RANSAC in Alg.~\ref{alg:pome}, Line~\ref{alg:pome:line:rel_rotation}).

We add a new connection $(j, k)$ to the view-graph if there are sufficient matches $\theta_{\text{matches}}$ ($=100$ in our implementation) after RANSAC (Alg.~\ref{alg:pome}, Line~\ref{alg:pome:upgade_vg}).

\subsection{Local View-Graph} \label{sec:local_vg}

\begin{figure}
	\centering
	{\includegraphics[width=0.47\textwidth]{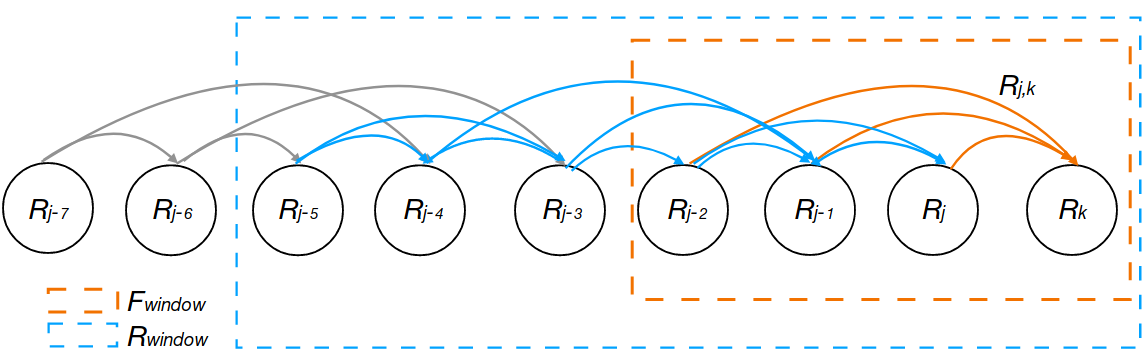}} \hspace{3em}
	\caption{The local view-graph $\cG'$. The nodes out of the $\cR_{\text{window}}$ remain constant within  our rotation averaging formulation.}\label{fig:local_view_graph}\vspace{-5mm}
\end{figure}

We estimate absolute camera orientations over a local sub-graph $\cG' = (\cV', \cE')$ of the view-graph $\cG = (\cV, \cE)$ to incrementally solve for their camera orientations at each frame step (Alg.~\ref{alg:pome}, Line~\ref{alg:pome:line:sub_graph}).

$\cV'$ contains the neighbouring nodes (with orientations in $\cR_{\text{window}}$) to the current frame $I_k$ and the adjacent vertices in $\cV$ to them (i.e., nodes in $\cV$ connected to the neighbouring nodes by an edge in $\cE$). Fig.~\ref{fig:local_view_graph} depicts $\cG'$.





\subsection{Incremental Rotation Averaging} \label{sec:pipeline:IRA}
Incrementally updating camera orientations in $\cG$ by solving ``conventional'' rotation averaging~\eqref{eqn:rot_avg} over a sub-graph of $\cG$ (``local'' to the current view) will inevitably produce imprecise results even after correcting drift. This is because the solution in the sub-graph will be anchored to the full graph at a single node (the first node in the sub-graph) with has an equivalent effect in accuracy to chaining orientations as discussed in Sec.~\ref{sec:rel_work:rot_avg}. In other words, the accuracy of the estimations in the local graph will be strongly dependant of the previous estimate of the first orientation in the sub-graph. To mitigate this chaining effect on accuracy, we propose an alternative formulation which anchors the solution in the sub-graph not to one but many nodes in the view-graph. To this end, for a window $\mathcal{R}_\text{window}$ of the most recent absolute orientation in $\cG$, we build its local view-graph $\cG'=(\cV', \cE')$ as described in Sec.~\ref{sec:local_vg}. Thus, we anchor the solution of
\begin{align}\label{eq:rotavg_new}
\min_{\cR_{\text{window}}} \sum_{(j,k)\in \cE'} \rho (d( \bR_{j,k}, \bR_k \bR_j^T )),
\end{align}
to all previously computed absolute orientations adjacent to the orientations in the window, i.e., the rotation matrices in 
\begin{align}
\cR_{\text{c}} = \cV' \setminus \cR_{\text{window}}.
\end{align}

To solve~\eqref{eq:rotavg_new}, we present an extension of L1-IRLS~\cite{chatterjee2013efficient}. L1-IRLS optimises over the $SO(3)$ Lie group by iteratively solving a weighted least-squares problem at the tangent Euclidean space which minimises 
\begin{align}
    \left(\sqrt{\Phi} \mathbf{A} \Delta \Omega_{\cV} + \sqrt{\Phi} \Delta \Omega_{\cE}\right)^2,
\end{align}
where $\Phi$ is a matrix collecting the weights for the edges in $\cE$, $\mathbf{A}$ is a matrix encoding $\cG$, $\Delta \Omega_{\cV}$ is the vector variable associated to the absolute orientations, and $\Delta \Omega_{\cE}$ the vector associated to the relative orientations (refer to~\cite{chatterjee2017robust} for details). Instead, we minimises
\begin{align}
    \left(\sqrt{\Phi'} \mathbf{A}' \Delta \Omega_{\text{window}} + \sqrt{\Phi'} \Delta \Omega_{\cE'}\right)^2,
\end{align}
where now the vector variable $\Delta \Omega_{\text{window}}$ comprises only the absolute orientations in $\cR_{\text{window}}$ and $\Delta \Omega_{\cE'}$ only relative orientations in $\cE'$. $\Phi'$ and $\mathbf{A}'$ are the versions for $\Phi$ and $\mathbf{A}$ without the unrelated columns to $\Delta \Omega_{\text{window}}$.

\subsection{Loop-Closure} \label{sec:loop_closure}
Loop-closure is an integral component in modern V-SLAM system. Our system detects (Alg.~\ref{alg:pome}, Line~\ref{alg:pome:lc_detec}) and solves (Alg.~\ref{alg:pome}, Line~\ref{alg:pome:line:global_rot_avg}) loop-closure when the camera captures images of a pre-visited scene. 


Our loop-closure routine consists of two sub-routines: 1) loop candidates detection, and 2) loop candidates validation. We adopted ORB-SLAM2's loop candidates detection routine due to its superior runtime and robustness. In essence, our system maintains a database that stores the DBoW2 features of every processed frame. For every incoming frame, the database is being queried with the latest frame to identify matching candidates. 

For the second part of the loop-closure, we invoke our feature extraction to relative orientation computation routines (Alg.\ref{alg:pome}, Line~\ref{alg:pome:line:feature_extraction} to Line~\ref{alg:pome:line:rel_rotation}) and validate the legitimacy of the candidates based on the number of inlier feature matches $|\cC_{j,k}'|$. This process is similar than in ORB-SLAM2, but instead of computing the similarity $Sim(3)$ transform with two sets of 3D point clouds (seen in the loop-closing pair of frames), we compute the relative orientations, given 2D-2D feature correspondences in the loop closing pair of frames. Upon adding the loop-closure edges, we solve a system-wide rotation averaging problem (Alg.~\ref{alg:pome}, Line~\ref{alg:pome:line:global_rot_avg}).


\section{Experiments}
\label{sec:exp}
We first provide empirical results to demonstrate the quality of our orientation estimates. Then, we showcase two potential SLAM applications for our proposed method: the known rotation problem (Sec.~\ref{sec:KRot}), and pure rotation motion (Sec.~\ref{sec:rot-only}). Our system requires of only three hyperparameters: the size of $\cF_{\text{window}}$, the size of $\cR_{\text{window}}$, and $\theta_{\text{matches}}$. We set them to $|\cF_{\text{window}}|=4$ , $|\cR_{\text{window}}|=10$, and $\theta_{\text{matches}}=100$ for all the experiments.

We implemented our proposed method in C++ and MATLAB R2019b.  All the experiments were conducted on a standard PC with six Intel i5-8400 CPU @ 2.80 GHz cores. The relative orientation estimation module (Sec.~\ref{sec:pipeline:rel_rotation}) was adopted from the OpenGV~\cite{kneip2014opengv} library.



\subsection{Datasets and Evaluation Metrics}
\label{sec:exp:datasets_metrics}
KITTI odometry benchmarking~\cite{geiger2012we} is one of the most popular datasets in the VO/V-SLAM community. We use the 11 sequences (00-10) with provided camera pose ground truths. Besides, we also perform experiments on our own dataset - the Carpark dataset. This sequence was collected with a camera and an IMU sensor attached on a driving vehicle in an open car park. The sequence has 1600 images which involves four strong rotations chunks (lead by turning motion) which is challenging for monocular VO and V-SLAM systems.

We benchmark with standard evaluation metrics. The first metric is the \emph{average rotation error} (as proposed in~\cite{geiger2012we})
\begin{align}
    r_{\text{err}} := \frac{1}{|\cF|} \underset{(i,j)\in\cF}{\sum} \angle ((\hat{\bR}_i^T \hat{\bR}_{j})^T (\bR_i^T \bR_{j})),
\end{align}
where $\cF$ is a set of pair of frame indices $\{(i,j)\}$, where the distance between frames in each pair varies from 100m to 800m\footnote{See the official site \url{http://www.cvlibs.net/datasets/kitti/eval\_odometry.php}}, $\hat{\bR}$ and $\bR$ are the ground-truth and the estimated orientations, and $\angle(\cdot)$ returns the  angle of the rotation matrix.


Our second metric is the rotation component of the \emph{Relative Pose Error}  (RPE)~\cite{sturm2012benchmark} for which we used two variants over $n$ camera poses. For a pre-defined ``time-step'' $\Delta$, the first variant is the RMSE over the $m := n-\Delta$ angular residuals
\begin{align} \label{eqn:rel_rot_error}    E_{i} := \angle ((\hat{\bR}_i^T \hat{\bR}_{i+\triangle})^T (\bR_i^T \bR_{i+\triangle})).
\end{align}
More explicitly, 
\begin{align}
    \begin{split}\label{eqn:RPE1}
        \text{RMSE}(E_{1:n}, \triangle) &:= \left(\frac{1}{m} \sum_{i = 1}^m E_i^2 \right)^{1/2}.
    \end{split}
\end{align}




As recommended by~\cite{sturm2012benchmark}, $\triangle$ should be set to 1 for VO systems that consider only frame-to-frame accuracy. As such the error reflects the rotation drift between consecutive pair of frames. Meanwhile, for V-SLAM system that emphasis on the overall drifting as well, we use the second RPE variant. The second RPE variant iterates $\triangle$ over a set of numbers ranging the first frame to the last frame $n$, and
\begin{align}
    \begin{split}\label{eqn:RPEn}
        \text{RMSE}(E_{1:n}) &:= \frac{1}{n} \sum_{\triangle = 1}^n \text{RMSE}(E_{1:n}, \triangle)\\
    \end{split}
\end{align}
averages the root-mean-square-error over $n$.

Henceforth we refer Eq.~\eqref{eqn:RPE1} as RPE$_1$ and Eq.~\eqref{eqn:RPEn} as RPE$_n$. Note that RPE$_n$ is the toughest metric since it involves the furthest pair of frames among the three metrics.

\begin{figure}
	\centering
	\includegraphics[height=4.2cm]{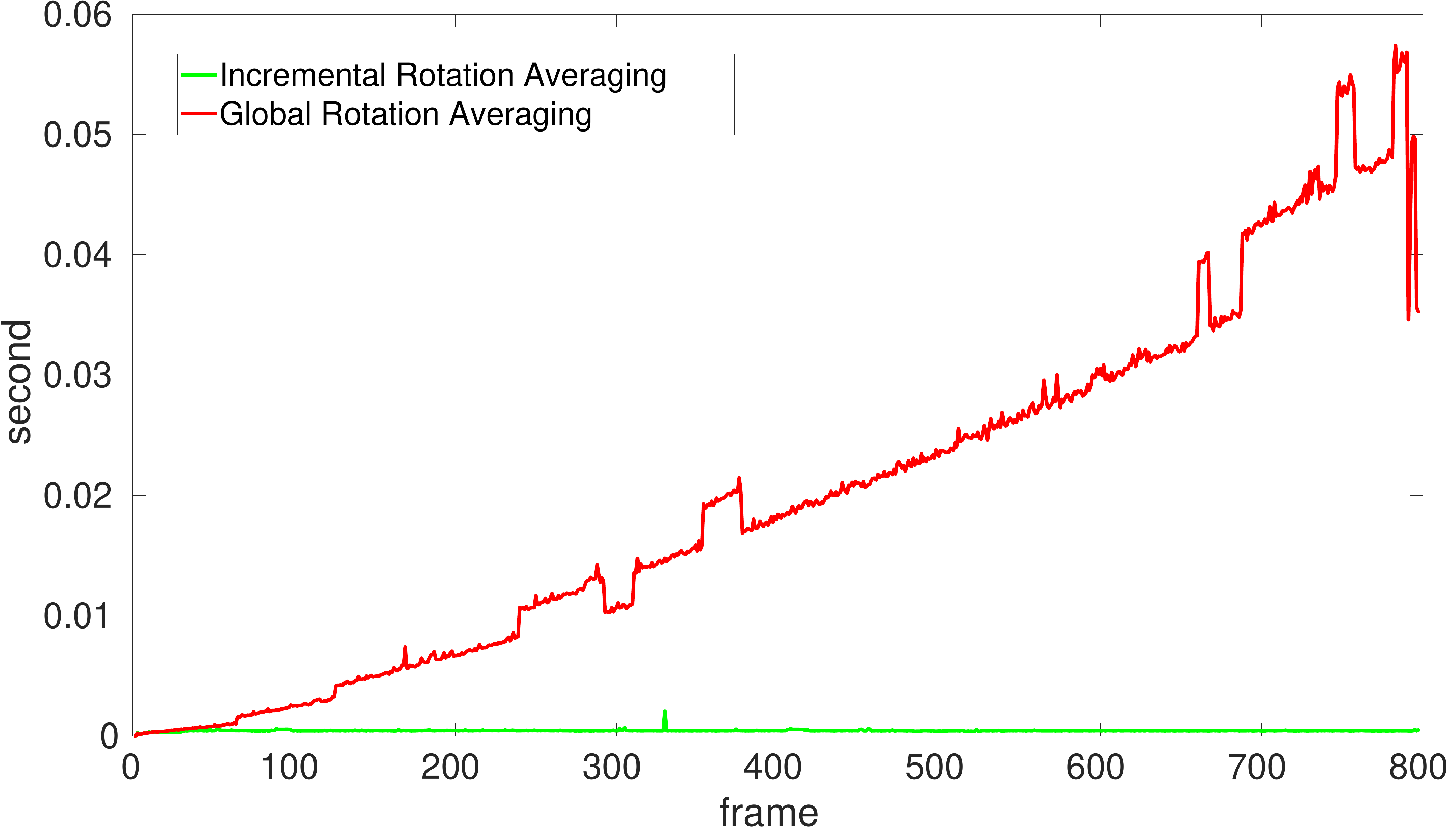} 
	\caption{Runtime comparison for rot. averaging in an incremental and global fashion on the KITTI 03 seq. Our incremental strategy exhibits constant runtime per frame (avg. 0.45ms).}\vspace{-5mm}\label{fig:runtime}
\end{figure}

\subsection{Ablation study of the incremental rotation averaging} \label{sec:exp:IRA}

\setlength{\tabcolsep}{4pt}
\begin{table}
	\begin{center}
		\caption{Ablation comparison for the incremental rotation averaging strategy. The significant RPE$_1$ error gap (lower the better) demonstrates the effectiveness of our incremental solution. Best results are in bold.}
		\label{tab:IRA}
		\begin{tabular}{cccccc}
		    \toprule
			 &  & \multicolumn{2}{c}{Baseline} & \multicolumn{2}{c}{Ours w/o loop closure}\\
			 \cmidrule(lr){3-4}\cmidrule(lr){5-6}
			 Dataset & Sequence & RPE$_1$ & RPE$_n$ & RPE$_1$ & RPE$_n$ \\
			  & & [deg] & [deg] & [deg] & [deg]\\
			\noalign{\smallskip}
			\hline
			\noalign{\smallskip}
			KITTI  & 00 & 0.36 & 8.67 &  \textbf{0.13} & \textbf{3.03}\\
			\noalign{\smallskip}
				   & 01 & 0.27 & 10.34 & \textbf{0.25} & \textbf{4.77}\\				   
			\noalign{\smallskip}
				  & 02 &  0.29 & 16.03 &\textbf{0.080} & \textbf{3.92}\\				   
  			\noalign{\smallskip}
				  & 03 & 0.28 & 5.47 &\textbf{0.053}& \textbf{0.65}\\				   
  			\noalign{\smallskip}
				  & 04 & 0.04 & 1.08&\textbf{0.034}& \textbf{0.50}\\				   
  			\noalign{\smallskip}
				  & 05 & 0.25 & 11.36&\textbf{0.053}& \textbf{2.27}\\				   
  			\noalign{\smallskip}
				  & 06 & 0.18 & 4.72&\textbf{0.049}& \textbf{1.41}\\				   
  			\noalign{\smallskip}
				  & 07 & 0.28 & 7.49& \textbf{0.065}& \textbf{1.04}\\				   
  			\noalign{\smallskip}
				  & 08 & 0.27 &9.21 &\textbf{0.055}& \textbf{3.18}\\				   
  			\noalign{\smallskip}
				  & 09 & 0.28 & 9.85&\textbf{0.051}& \textbf{1.37}\\				   
  			\noalign{\smallskip}
				  & 10 & 0.38 & 13.25&\textbf{0.061}& \textbf{2.30}\\				   
   			\noalign{\smallskip}
   			\hline
   			\noalign{\smallskip}
   			Carpark & & 0.95 & 3.48 &\textbf{0.14} & \textbf{1.05}\\				   
			\hline
		\end{tabular}
	\end{center}
\end{table}
\setlength{\tabcolsep}{1.4pt}

We first evaluate the effectiveness of our incremental rotation averaging component by comparing against a baseline without this component. Instead, the baseline estimates the absolute orientations by chaining relative orientations (Eq.~\eqref{eqn:rel_rot}). We deactivate the the loop-closure component (Lines~\ref{alg:lc_begin}-\ref{alg:lc_end} in Alg.~\ref{alg:pome}) in both our method and the baseline. 

As explained earlier, RPE$_1$ is the relevant comparing metric for this experiment since our method can be regarded as a pure VO system when loop-closure is deactivated. The results in Tab.~\ref{tab:IRA} show that the incremental rotation averaging strategy has a substantial impact on our system, significantly outperforming the baseline on every sequence. On average, we outperformed RPE$_1$  in 0.181 degrees over the KITTI dataset, and in 0.81 degrees on the Carpark dataset. We also reported  RPE$_n$ values to compare against the results with loop-closure reported in Tab.~\ref{tab:sota_benchmark}. Note that the performance of our method with the loop-closure module activated improves significantly on all the sequence with loops (e.g., KITTI sequences 00, 02, 05, 06, 07, 09, and in the Carpark dataset).

A major advantage of our incremental rotation strategy is scalability. By fixing the absolute orientations to estimate to a constant $|\cR_{\text{window}}|$ at every frame step, solving our proposed formulations (Eq.~\eqref{eq:rotavg_new}) takes constant runtime. To demonstrate the speed gain, we compare against solving ``conventional'' rotation averaging~\eqref{eqn:rot_avg} over the entire view-graph at every frame step. Fig.~\ref{fig:runtime} plots the runtimes for the comparison. The runtime for solving rotation averaging over the entire view-graph grows as the number of frame processed increases. Meanwhile, our incremental strategy maintains a constantly low processing time (0.45ms on average) at every frame step.


\subsection{Accuracy of Camera Orientation} \label{sec:exp:accu}
\begin{table}
	\begin{center}
		\caption{Comparison of our proposed method against state-of-the-art monocular VO and V-SLAM system. The best results are in bold; second best results are underlined.}
		\label{tab:sota_benchmark}
		\begin{tabular}{ccccccccccc}
		\toprule
			& & \multicolumn{3}{c}{DF-VO\cite{zhan2019visual}}&
			\multicolumn{3}{c}{ORB-SLAM2\cite{mur2017orb}}&
			\multicolumn{3}{c}{Ours}\\
			\cmidrule(lr){3-5}\cmidrule(lr){6-8} \cmidrule(lr){9-11}
			Datasets &  Seq. & $r_{\text{err}}$ & RPE$_{1}$ & RPE$_{n}$  & $r_{\text{err}}$ & RPE$_{1}$ & RPE$_{n}$  & $r_{\text{err}}$ & RPE$_{1}$ & RPE$_{n}$  \\
			& & {\tiny [deg/100m]}  & {\tiny [deg]} & {\tiny [deg]} & {\tiny [deg/100m]} & {\tiny [deg]} & {\tiny [deg]} &{\tiny [deg/100m]} & {\tiny [deg]} & {\tiny [deg]} \\
			\noalign{\smallskip}
			\hline
			\noalign{\smallskip}
			KITTI  & 00 & 0.58 & \textbf{0.12} & 4.06 & \textbf{0.33} & 0.38 & \underline{2.27}  & \underline{0.41} & \underline{0.13} & \textbf{1.88}\\
			\noalign{\smallskip}
			& 01 & 17.04 & 9.95 & 124.33 & \textbf{0.42} & \underline{0.70} & \underline{6.10}  & \underline{0.86}  & \textbf{0.26} & \textbf{5.53}\\				   
			\noalign{\smallskip}
			& 02 & 0.52 & \textbf{0.072} & 4.37 &\textbf{0.30} & 0.17 & \textbf{2.47}  & \underline{0.40} & \underline{0.081} & \underline{3.06}\\
			\noalign{\smallskip}
			& 03 & 0.39 & \textbf{0.049} & 0.95 & \textbf{0.21} & 0.062 & \textbf{0.35}   & \underline{0.30} & \underline{0.052} & \underline{0.66}\\
			\noalign{\smallskip}
			& 04 & \textbf{0.25} & \underline{0.036} & \underline{0.45} & \underline{0.31} & 0.063 & \textbf{0.33}  & 0.33  & \textbf{0.035} & 0.54\\
			\noalign{\smallskip}
			& 05 & \underline{0.30}& \textbf{0.048} & 1.62 & \textbf{0.26}& 0.14 & \textbf{1.16}  & 0.34 & \underline{0.055} & \underline{1.27}\\
			\noalign{\smallskip}
			& 06 & 0.30& \textbf{0.037} & \textbf{0.89} & \textbf{0.24} & 0.11 & 1.07  & \underline{0.27} & \underline{0.053} & \underline{0.94}\\
			\noalign{\smallskip}
			& 07 & \textbf{0.27} & \textbf{0.037} & \textbf{0.78} &0.49 & 0.081 & 1.12  & \underline{0.41}  & \underline{0.075} & \underline{0.99}\\
			\noalign{\smallskip}
			& 08 & \underline{0.32} & \textbf{0.050} & \underline{2.24} & \textbf{0.32} & 0.090 & \textbf{1.83} & 0.39 & \underline{0.055} & 3.38\\
			\noalign{\smallskip}
			& 09 & \textbf{0.29} & \textbf{0.045} & 1.70 & \underline{0.31} & 0.11 & \underline{1.47}  & \underline{0.31} & \underline{0.050} & \textbf{1.08}\\
			\noalign{\smallskip}
			& 10 & \textbf{0.37} & \textbf{0.053} & \textbf{1.36} & \underline{0.38} & 0.13 & 2.75 & 0.49 & \underline{0.062} & \underline{2.57}\\
			\noalign{\smallskip}
			\hline
			\noalign{\smallskip}
			Carpark & & - & - & - & \underline{0.32}& \textbf{0.14} & \textbf{0.41}& \textbf{0.27} & \textbf{0.14}& \underline{0.48}\\
			\bottomrule
		\end{tabular}
	\end{center}
\end{table}

We compare against the state-of-the-art VO and V-SLAM systems, namely DF-VO~\cite{zhan2019visual} and ORB-SLAM2\cite{mur2017orb}, and reported results in Tab.~\ref{tab:sota_benchmark}. As for the results of the DF-VO method, we downloaded DF-VO's estimates (KITTI) from their official github repo\footnote{https://github.com/Huangying-Zhan/DF-VO}. As of ORB-SLAM2, we ran it on all tested sequences five times and report the median results. Similarly, we report the median of five different runs of our proposed method.


Our system achieves competitive performance to the state-of-the-art on both the KITTI and the Carpark datasets over all metrics. The RPE$_1$ evaluation shows that DF-VO's frame-to-frame drift is the best among the three methods. ORB-SLAM2 tends to have a higher frame-to-frame error due to its frame management scheme that skips consecutive frames with similar appearances. In terms of RPE$_n$, our method outperforms ORB-SLAM2 in 5 KITTI sequences (00, 01, 06, 07, and 09). As mentioned earlier, comparing RPE$_n$ with DF-VO is unfair due to the lack of loop-closing ability.

\begin{figure}
	\centering
	\includegraphics[width=7.5cm]{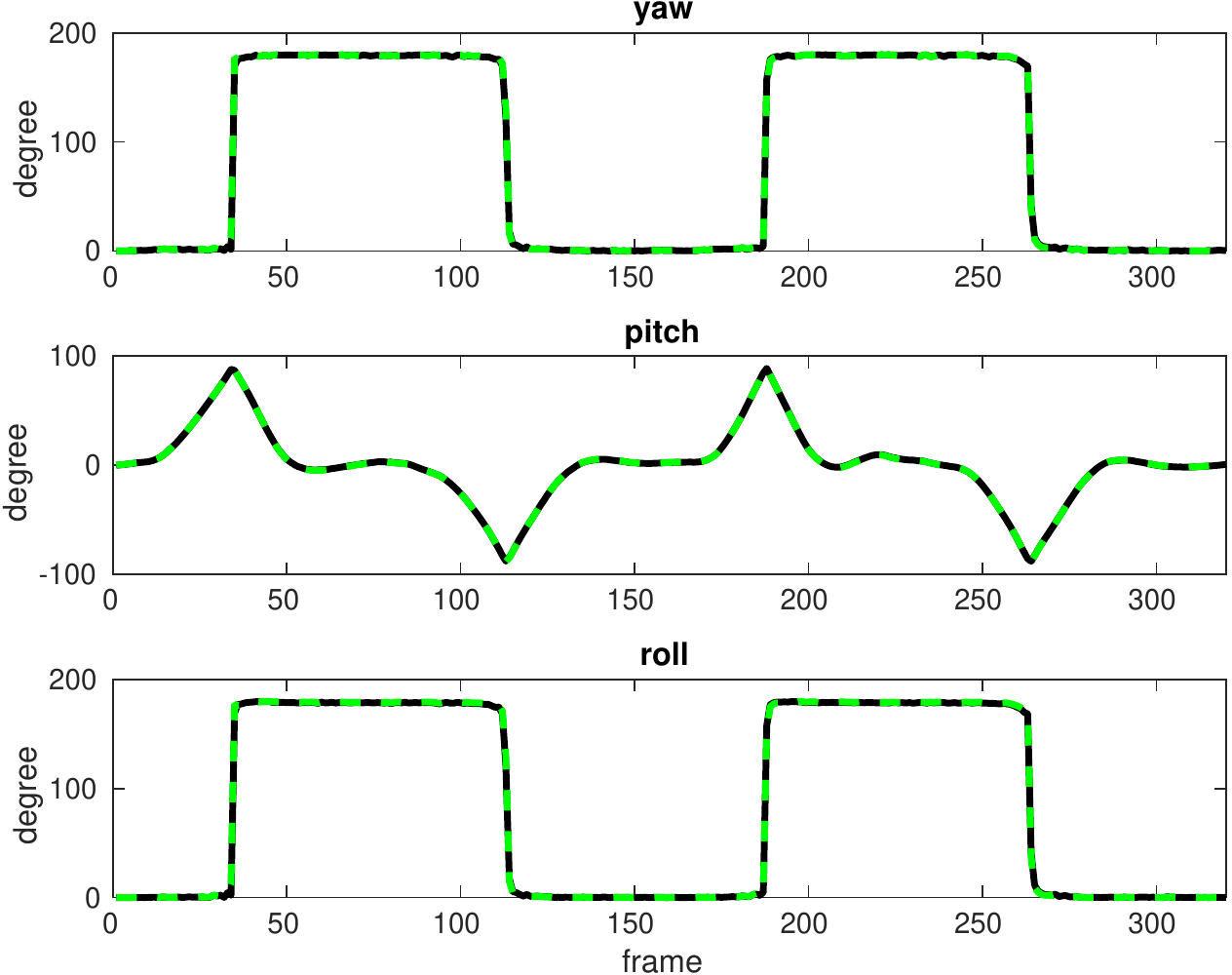} 
	\caption{The yaw, pitch, roll comparisons between ground-truth (black solid line) and our proposed method (green dash line) on the Carpark dataset. The dataset has four strong rotation chuncks (see ``pitch'' plot) caused by vehicle in turning motion. Our method manage to produce high accuracy orientation estimates from 2D-2D feature correspondences alone.}\label{fig:mt_rpy}\vspace{-10mm}
\end{figure}

To provide a better insight of our orientation estimates, we convert the rotation matrix into the Euler angles (yaw, pitch, and roll) and plot it against the ground-truth of the Carpark dataset in Fig.~\ref{fig:mt_rpy}. We observe that the orientation variation throughout the sequence is significant and our estimates closely align with the ground-truths.

\subsection{Application A - The Known Rotation Problem} \label{sec:KRot}

\begin{figure*}
	\centering
	\subfloat[KITTI seq. 07]{\includegraphics[height=5cm]{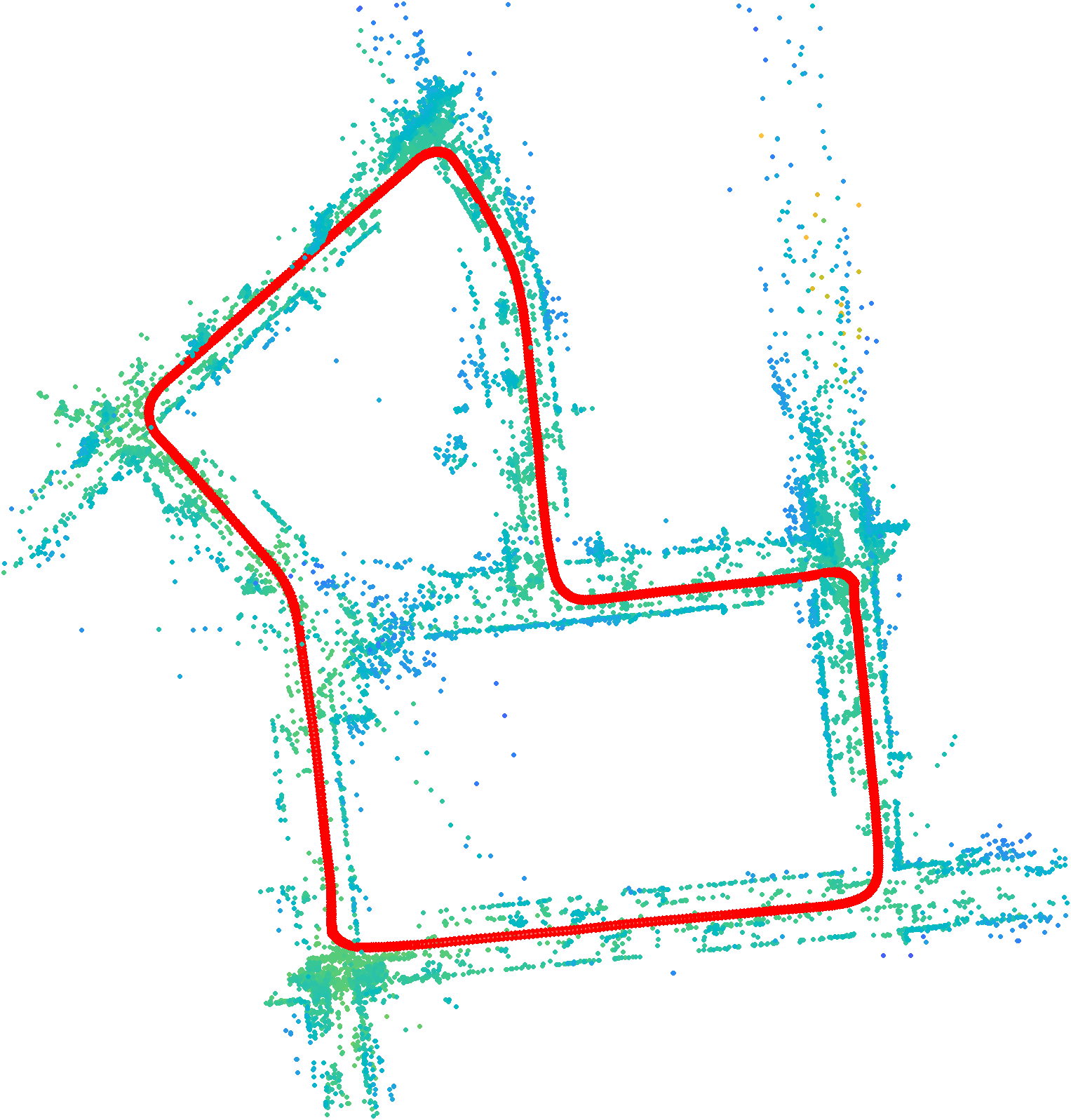} \label{fig:kt_07}}\hspace{3em}
	\subfloat[Carpark dataset]{\includegraphics[height=5cm]{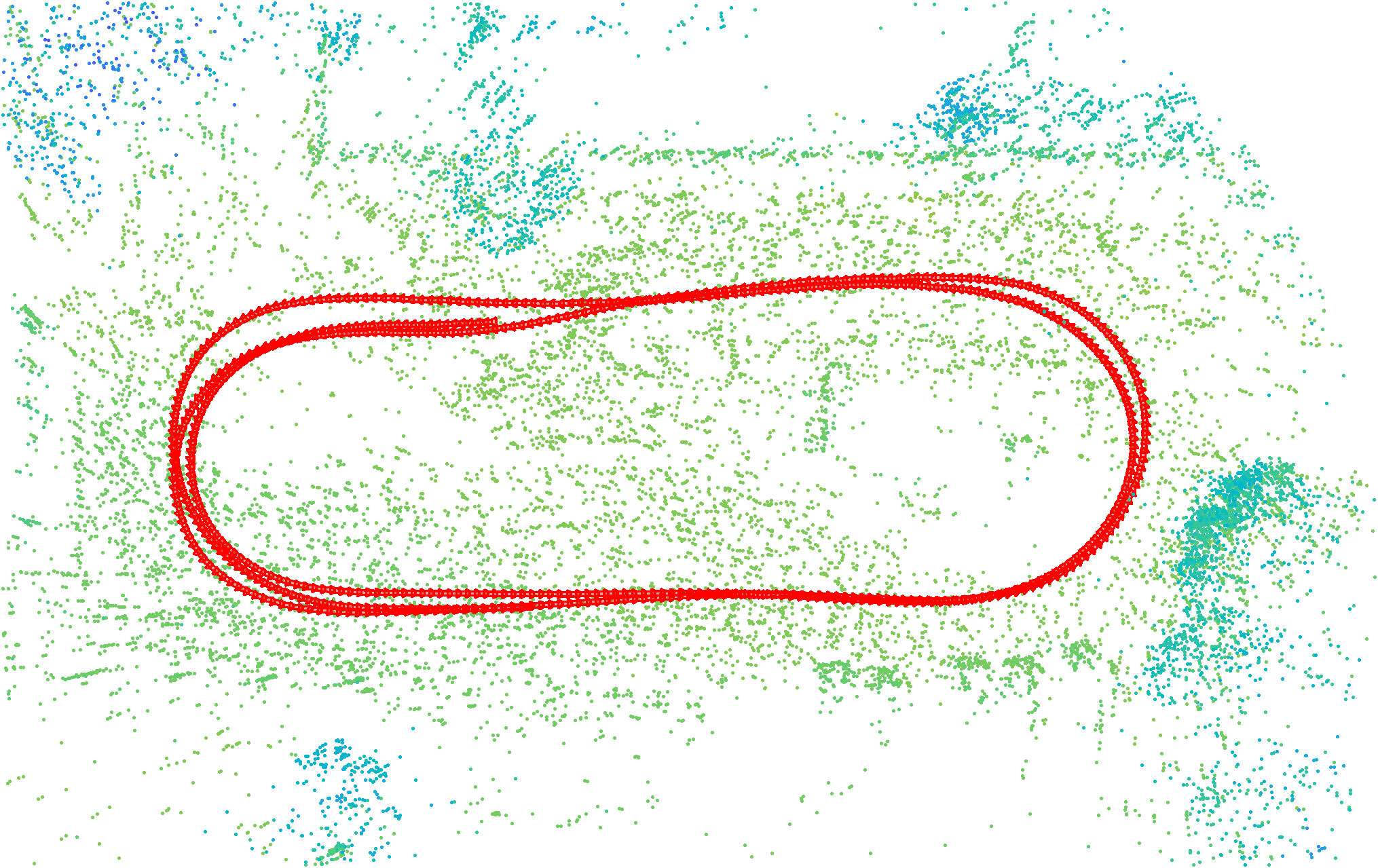} \label{fig:mt}}\\
	\caption{The SLAM outputs of Application A (Sec.~\ref{sec:KRot}) for (a) the the KITTI dataset seq. 07, and (b) the Carpark dataset.}\label{fig:slam_outputs}\vspace{-1mm}
\end{figure*}

An application of our system is to provide the camera orientations $\{\bR_j\}$ to then estimate the camera positions $\{\bt_j\}$ and the scene coordinates $\{\bX_i\}$ by solving the known rotation problem~\cite{kahl2008multiple}
\begin{align*} \label{eqn:krot_opt}
\begin{aligned} 
&\min_{ \{\bt_j\}, \{\bX_i\} } \max_{i,j} \| \bx_{i,j} - f(\bX_i | \bR_j, \bt_j) \| \\
&  \text{s.t.} \quad  \bR_j^{(3)} \bX_i + \bt_j^{(3)} > 0 \quad  \forall (i,j) \in \cE,
\end{aligned}
\refstepcounter{equation}\tag{\theequation}
\end{align*} 
where $f(\cdot)$ projects scene points into the image plane.

The objective of the known rotation problem is quasi-convex allowing global optimisation. We used the fast Res-Int solver~\cite{zhang2018fast} to solve~\eqref{eqn:krot_opt}. Finally we refine all the variables ($\{\bR_j\}, \{\bt_j\}, \{\bX_i\}$) with BA. Fig.~\ref{fig:slam_outputs} displays the results.



\subsection{Application B - Pure Rotation Motion} \label{sec:rot-only}

\begin{figure*}
	\def\figh{3.5cm}
	\begin{tabularx}{\textwidth}{c|C{.35}|C{.3}|C{.3}|C{.3}}
		& {\small Google Map} & {\small ORB-SLAM2} & {\small OpenVSLAM} & {\small Ours}\\
		
		\hline
		\rotatebox[origin=l]{90}{\small ~~~~~~~~~~ \emph{seq1}}&
		\includegraphics[height=2.5cm]{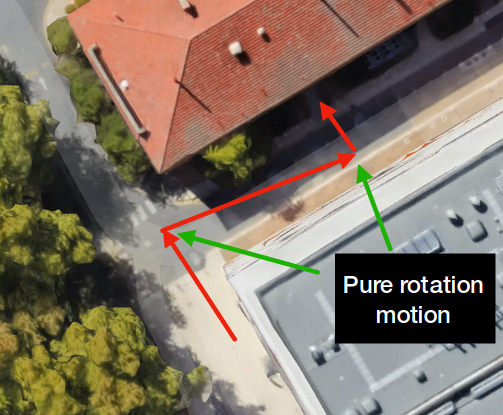} &
		\includegraphics[height=2.3cm]{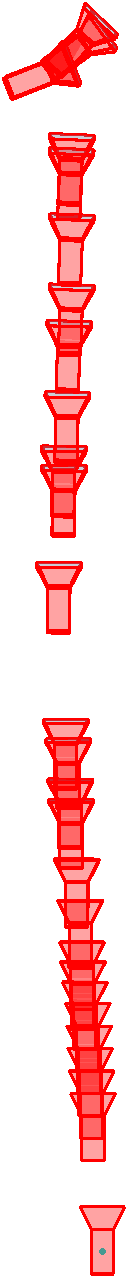} &
		\includegraphics[height=2.3cm]{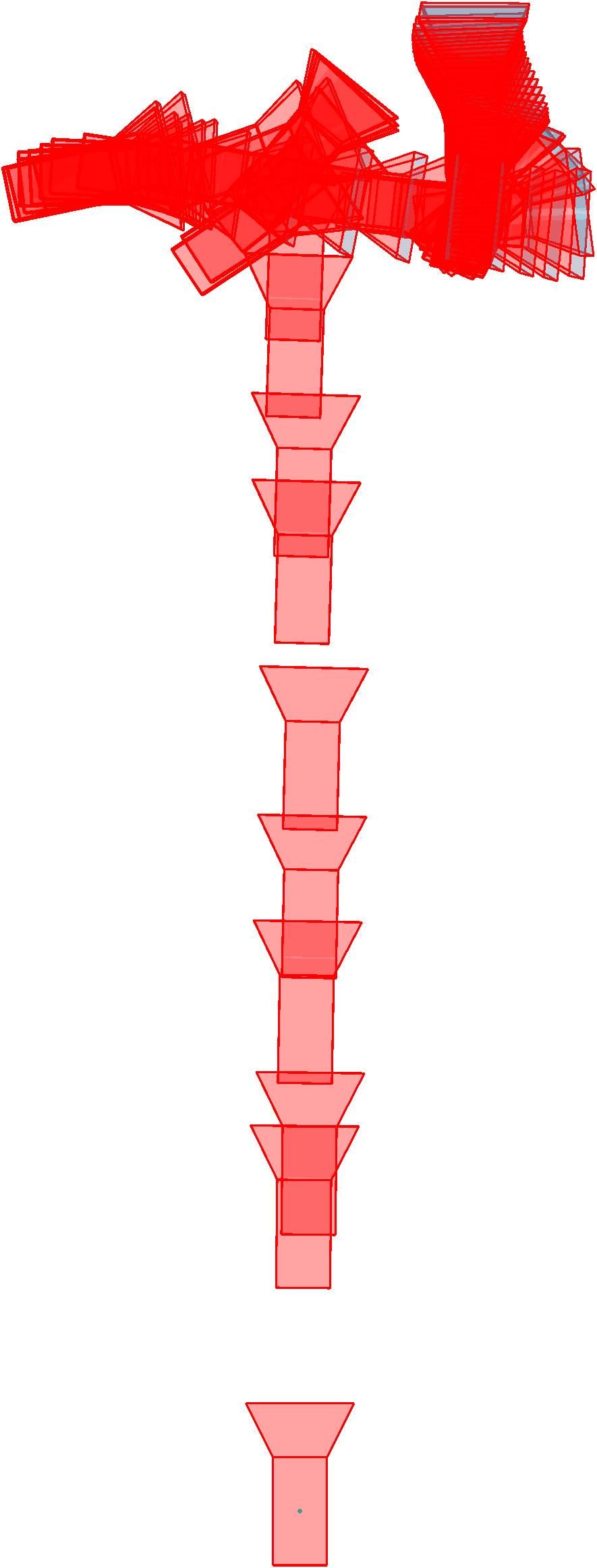} &
		\includegraphics[width=2.5cm]{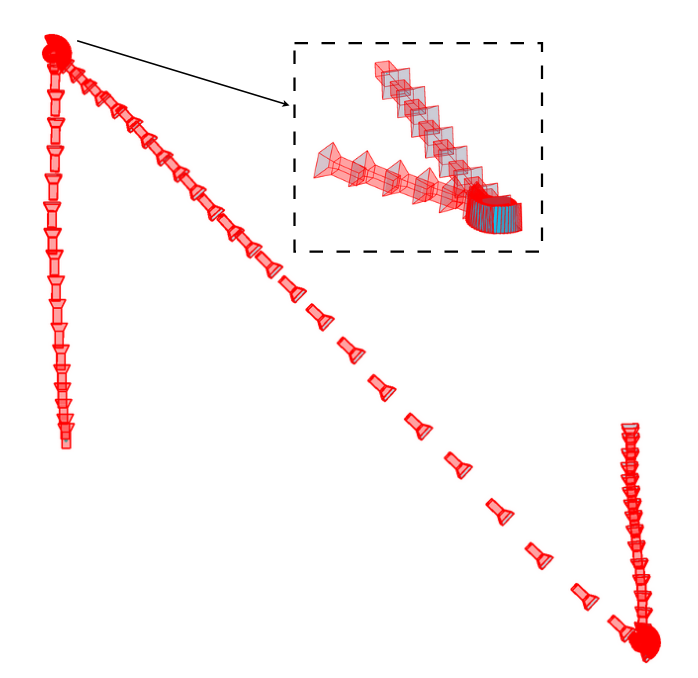} \\
		\hline
		
		\rotatebox[origin=l]{90}{\small ~~~~~~~~~~ \emph{seq2}}&
		\includegraphics[height=2.5cm]{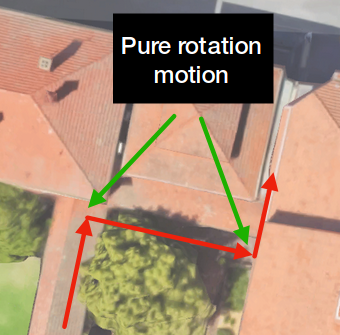} &
		\includegraphics[height=2.3cm]{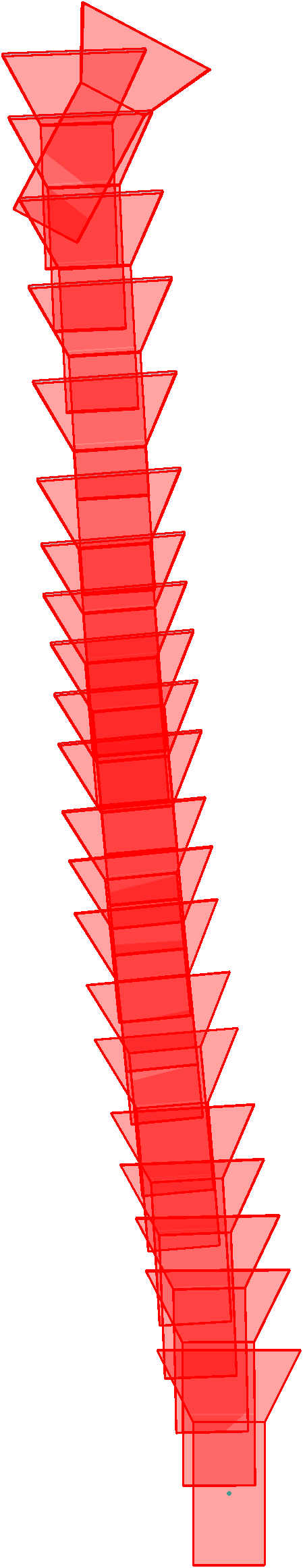} &
		\includegraphics[height=2.3cm]{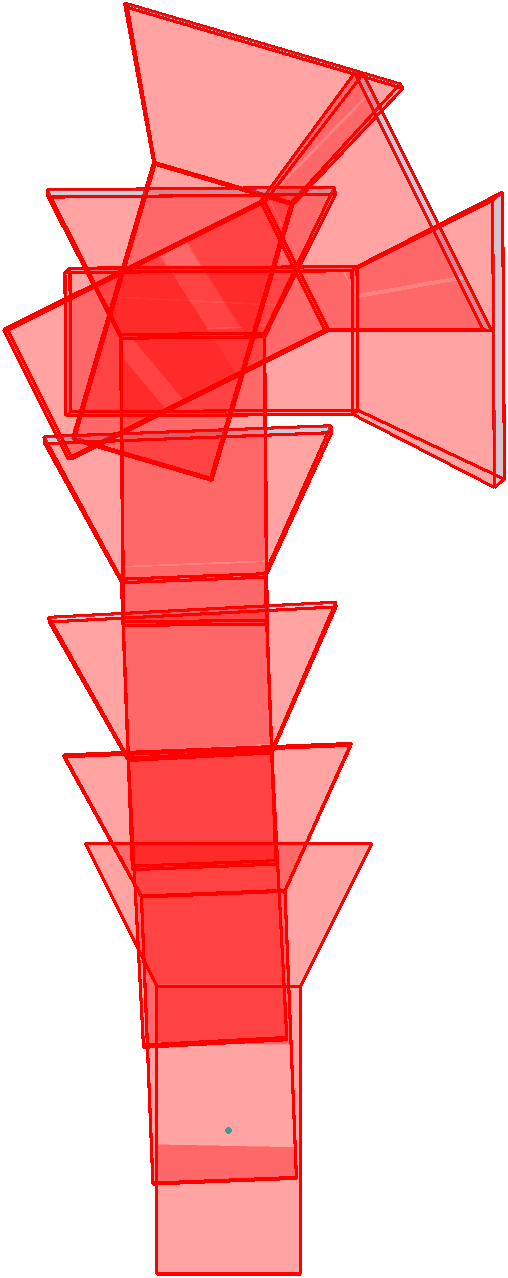} &
		\includegraphics[height=2.5cm]{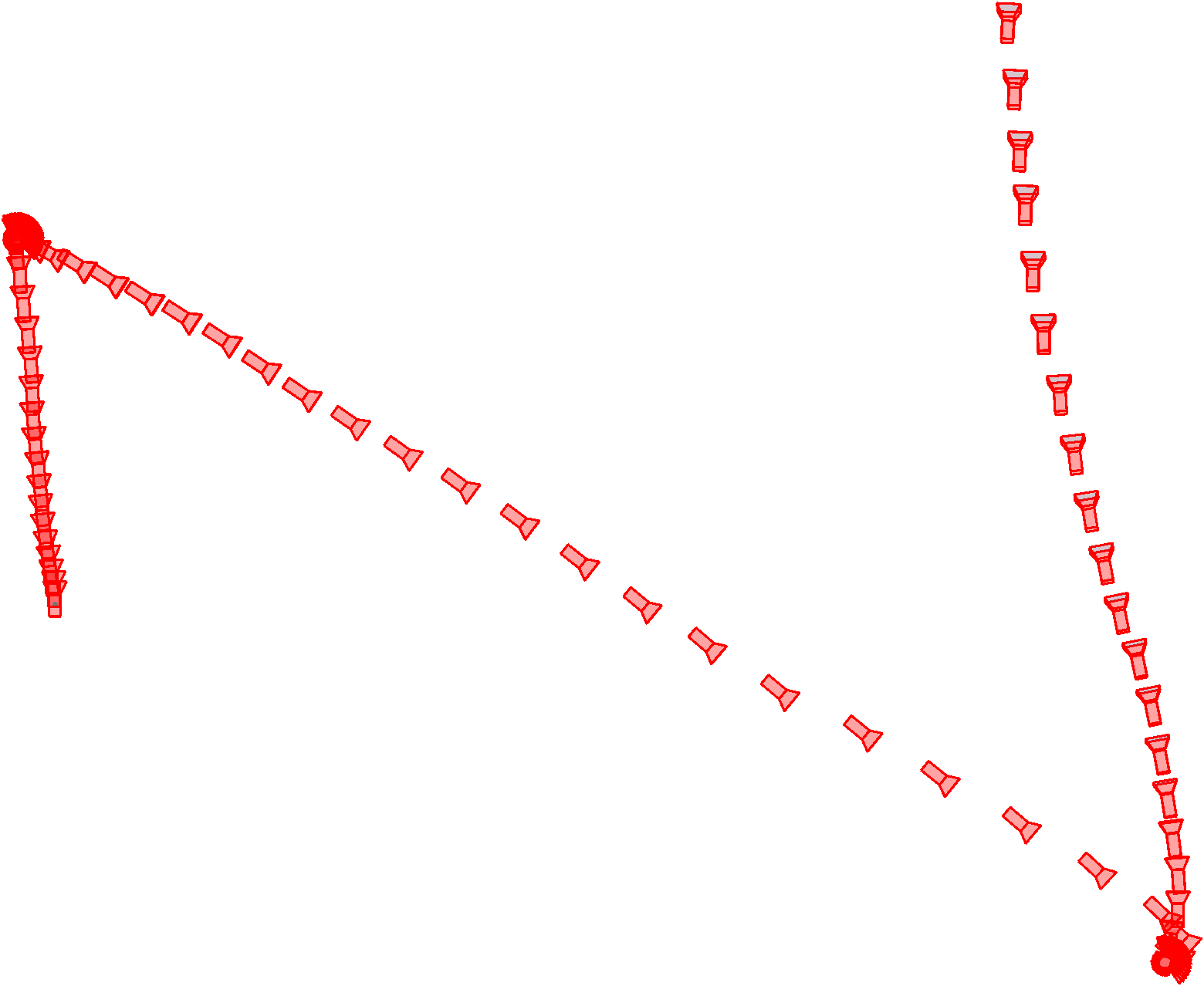} \\
		
		\hline
		\rotatebox[origin=l]{90}{\small ~~~~~~~~~~ \emph{seq3}}&
		\includegraphics[height=2.5cm]{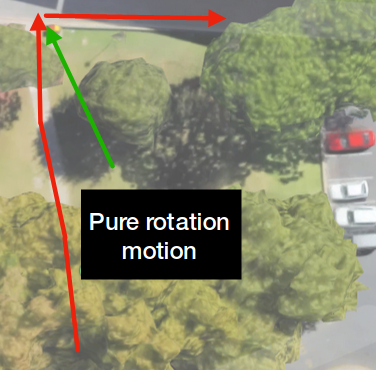} &
		\includegraphics[height=2.3cm]{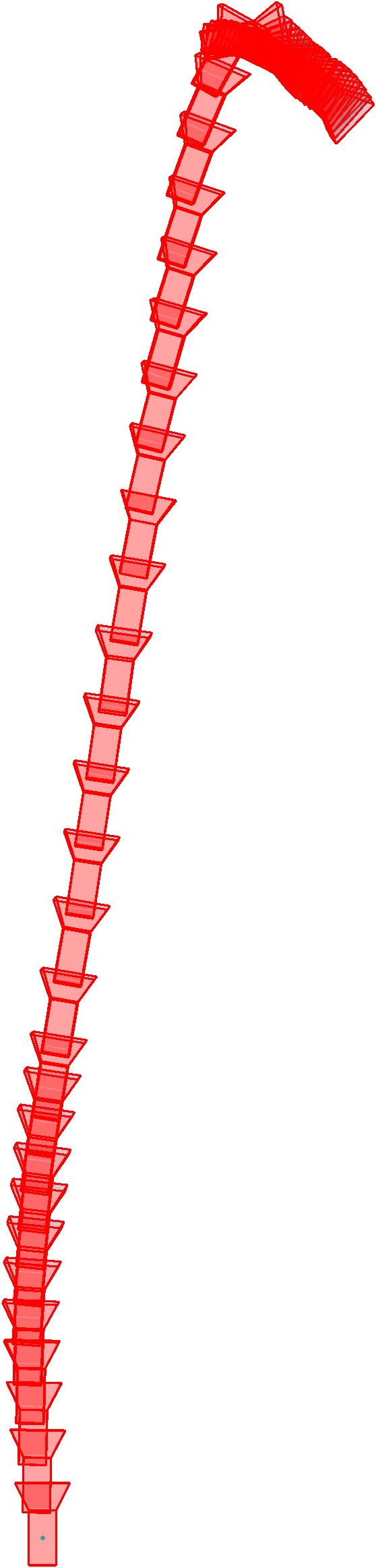} &
		\includegraphics[height=2.3cm]{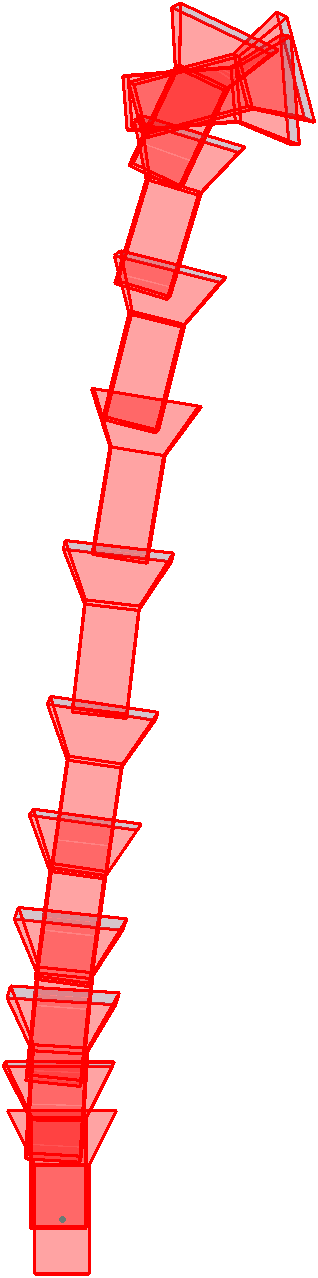} &
		\includegraphics[height=2.5cm]{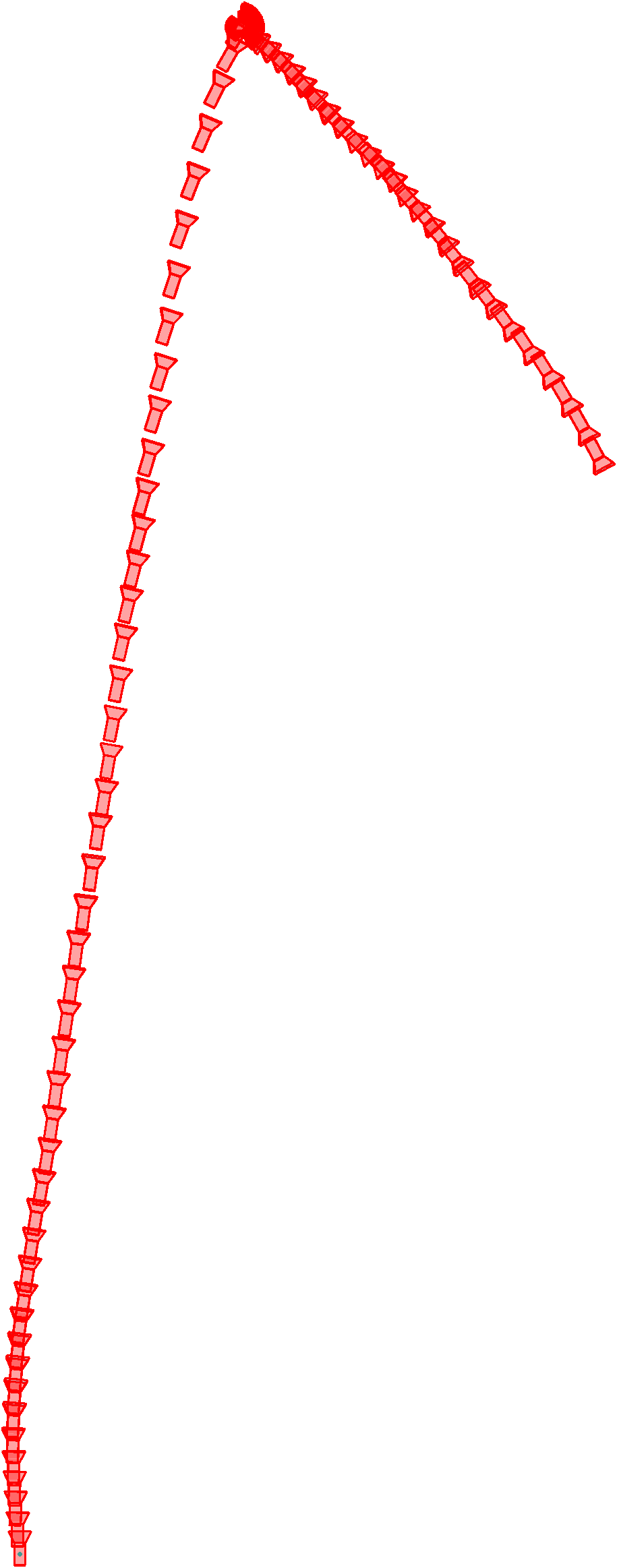} \\
	\end{tabularx}	
	\caption{Camera trajectories (sampled for better visualisation) for ORB-SLAM2, OpenVSLAM and our method over three sequences that contains  rotation-only motions. The first column shows the approx. trajectories (captured from GoogleMap).}\vspace{-5mm}
	\label{fig:pr_example}
\end{figure*}

Modern monocular visual SLAM systems (e.g., ORB-SLAM2, OpenVSLAM~\cite{sumikura2019openvslam}) utilise BA as the core engine to jointly estimate the camera poses and scene points. One drawback of such approach is that BA cannot handle rotation-only motion as the 3D scene points can not be triangulated\footnote{We are referring to the scenarios where the pure rotation motion is long enough where the camera is turned to a new scene and observed a new scene where none of the features were triangulated previously.}. We captured three sequences where our mobile phone is attached to a gimbal and rotate around a principal axis to simulate such a motion. Fig.~\ref{fig:pr_example} depicts the resulting poses of ORB-SLAM2, OpenVSLAM (latest implementation of ORB-SLAM2-like monocular SLAM pipeline), and our method on these sequences. Note that the commercial based mobile phone camera is not precisely calibrated\footnote{We calibrated with a checkerboard.} which causes the drift in the estimated trajectory.

Most of the examples in Fig.~\ref{fig:pr_example} (except (seq1, ORB-SLAM2) and (seq3, OpenVSLAM)) show that the tracking of the camera pose stopped during long pure rotation motion part of the sequence.  Note that both methods employ a scene point selection heuristic (based on reprojection error) to decide if the newly triangulated point should be added to the existing local map. During pure rotation motion, most of the triangulated points were successfully trimmed away by the heuristic and eventually went into ``relocalisation mode'' when their local map run out of 3D points. However, in some cases where the heuristic fails to reject these points and continue to track the camera poses, the scale of the camera trajectory after the pure rotation motion became inaccurate as for (seq3, ORB-SLAM2), and (seq1, OpenVSLAM).


Since our method detaches the orientation component of the camera poses from the joint estimation framework, the estimation of the camera orientation remains unaffected during pure rotation motion. We skip the known rotation problem and the BA routines (as explained in Sec.~\ref{sec:KRot}) in the rotation-only frames and update the camera orientations with the outputs of our proposed method. Meanwhile, we fix the camera positions during those motions. 

\setlength{\tabcolsep}{1.4pt}

\section{Conclusion}
We presented a novel monocular rotational odometry system that is capable of producing accurate camera orientation estimates. The key aspects of proposed method are 1) its motion robustness, 2) a new constant time incremental rotation averaging solver, and 3) the ability to perform loop closure. Our proposed method achieve state-of-the-art accuracy on real-world datasets. Lastly, two potential applications of our method were demonstrated.




%

%


\bibliographystyle{IEEEtran}
\bibliography{egbib}

\end{document}